\documentclass[conference]{IEEEtran}
\IEEEoverridecommandlockouts
\usepackage{cite}

\usepackage{tikz} 
\usepackage{listings}
\usepackage{enumitem}
\usepackage{graphicx}
\usepackage{multicol, latexsym}
\usepackage{blindtext}
\usepackage{subcaption}
\usepackage{caption}
\usepackage{longtable}

\usepackage{adjustbox}  

\usepackage{csquotes}
\usepackage{amsfonts}
\usepackage{amsmath}
\usepackage{amsthm}
\usepackage{amssymb}
\usepackage{algorithm}
\usepackage{tabularx}
\usepackage{capt-of,lipsum} 

\usepackage{listings}
\usepackage{hyperref}
\usepackage{comment}
\usepackage{diagbox}
\usepackage{pdflscape}
\usepackage{booktabs}
\usepackage{makecell}
\usepackage{balance}

\usepackage{array}

\usepackage{dirtytalk}
\usepackage{lipsum}
\usepackage{url}

\usepackage{mathtools, nccmath}

\usepackage{siunitx}
\DeclareSIUnit[number-unit-product = { }] \dBm{dBm}

\usepackage{algpseudocode}
\usepackage{algorithm}

\usepackage{listing-styles}

\algnewcommand\algorithmicforeach{\textbf{for each}}
\algdef{S}[FOR]{ForEach}[1]{\algorithmicforeach\ #1\ \algorithmicdo}
\usepackage[most]{tcolorbox}
\usepackage{multirow}

\definecolor{myblue}{rgb}{0.09,0.20,0.34}
\definecolor{mygreen}{rgb}{0,0.6,0}
\definecolor{mygray}{rgb}{0.98,0.98,0.98}
\definecolor{myorange}{rgb}{0.92,0.49,0.34}
\definecolor{mywhite}{rgb}{1.0,1.0,1.0}

\definecolor{NMR}{RGB}{255,255,86}
\definecolor{MRA}{RGB}{255,231,27}
\definecolor{MRD}{RGB}{178,178,178}
\definecolor{MRIP}{RGB}{188,172,0}
\definecolor{MRS}{RGB}{161,207,106}
\definecolor{NA}{RGB}{228,60,52}
\definecolor{AI}{RGB}{255,255,86}
\definecolor{RMR}{RGB}{162,4,21}
\definecolor{RMD}{RGB}{178,178,178}
\definecolor{RMA}{RGB}{255,231,27}
\definecolor{RMIP}{RGB}{188,172,0}
\definecolor{RMS}{RGB}{161,207,106}

\newcommand{\ie}{\textit{i}.\textit{e}.,\ }
\newcommand{\eg}{\textit{e}.\textit{g}.,\ }

\newsavebox{\mybox}

\usepackage{pifont}

\usepackage{censor}
\usepackage{wasysym}

\usepackage[scaled]{helvet}
\usepackage[T1]{fontenc}
\usepackage{helvet}

\usepackage{geometry}
\begin{document}

\title{Towards Interpretable Depression Detection:\\Linking Acoustic Features to DSM-5 Indicators}


\DeclareRobustCommand*{\IEEEauthorrefmark}[1]{%
  \raisebox{0pt}[0pt][0pt]{\textsuperscript{\footnotesize #1}}%
}

\author{\IEEEauthorblockN{Jonas Länzlinger\IEEEauthorrefmark{1},
Katharina O.E. Müller\IEEEauthorrefmark{2},
Burkhard Stiller\IEEEauthorrefmark{2},
Bruno Rodrigues\IEEEauthorrefmark{1}
}

\IEEEauthorblockA{\IEEEauthorrefmark{1}Embedded Sensing Group ESG, School of Computer Science SCS, 
University of St. Gallen HSG, Switzerland\\
} 
\IEEEauthorblockA{\IEEEauthorrefmark{2}Communication Systems Group CSG, Department of Informatics IfI, 
University of Zurich UZH, Switzerland\\
}
E-mail:jonas.laenzlinger@student.unisg.ch¦[stiller,mueller]@ifi.uzh.ch¦bruno.rodrigues@unisg.ch\\\\
\textit{This paper has been accepted at IEEE PerCom 2026 as a Work-in-Progress (WiP) paper.}
}


\maketitle

\begin{abstract}
Depression affects millions worldwide, yet diagnosis relies on subjective self-reports that may miss authentic behavior. This paper presents an approach linking speech acoustics to DSM-5 depressive-behavior indicators through a transparent Linkage Framework. Unlike black-box models, the framework explicitly maps acoustic features (pitch variability, pauses, speech tempo) to clinical indicators, enabling interpretable, indicator-level outputs. The system runs locally on commodity hardware (HW) to preserve privacy. Preliminary evaluation on DAIC-WOZ shows directionally consistent associations between acoustic features and DSM-5 indicators for psychomotor change and concentration difficulty, supporting the design rationale. Future work will validate on longitudinal datasets and extend multimodal integration while maintaining edge constraints.
\end{abstract}


\section{Introduction}
\label{sec:introduction}

Depression is one of the most prevalent mental health disorders, with rates rising across all age groups \cite{who2023depression, Remes2021DeterminantsDepression}. The consequences are especially severe for children and adolescents, where early symptoms are often overlooked \cite{ahrq2022child}. While clinical diagnosis relies on DSM-5 criteria and self-report questionnaires like PHQ-9 \cite{americanpsychiatricassociation2022dsm5, kroenke2001phq9}, these methods are susceptible to recall errors and subjectivity \cite{baumeister2007behavior}. As access to mental health support remains limited \cite{who2023depression}, innovative detection methods become critical, yet current systems face two barriers: \emph{interpretability} and \emph{privacy}.

\textbf{The interpretability gap.} Deep learning models achieve high accuracy, but their outputs (\textit{e.g.,} a single probability score) provide no insight into \emph{which symptoms} are elevated or \emph{why}. Clinicians cannot integrate such outputs into workflows requiring symptom-level reasoning aligned with DSM-5~\cite{fried2025depressionsumscores}. A transparent mapping between signals and clinical indicators is essential.

\textbf{The privacy imperative.} Speech data is inherently sensitive in household contexts. Cloud offloading raises major risks: recordings reveal identity and health status. Local processing ensures raw audio never leaves the home, which is critical for longitudinal monitoring.

Although combining speech, text, physiology, and behavior yields the most reliable assessments \cite{zhang2024multimodal,jahan2022multimodality}, integrating heterogeneous signals on constrained household devices remains impractical. Therefore, our approach concentrates on acoustic data from speech: a modality that carries clinically validated cues of depression and can be processed transparently and efficiently at the edge.

\textbf{Related work.} Depression detection has been explored across text, speech, vision, and mobile sensing. Text-based methods analyze social media posts using transformer embeddings \cite{sharma2020twitter,Cacheda2019earlyDD}, while vision approaches leverage facial expressivity and gaze \cite{cohn2009detecting,Min2023detectingDepressionVideoLogs}. Digital phenotyping uses passively collected signals from phones and wearables \cite{lin2020sensemood,shen2017dd}. For speech, prior work shows that acoustic cues (reduced pitch variability, slower speech, longer pauses) correlate with depressive states \cite{Low2020keyacousticfeatures,Donaghy2024VoiceBiomarkers}. However, most studies report aggregate accuracy without exposing which feature supports which DSM-5 indicator \cite{Yadav2022ADD}. Without precise feature-indicator mappings, outputs remain hard to interpret in clinical workflows.

It had been investigated whether acoustic features can be mapped to DSM-5 indicators in a transparent, resource-aware way. The challenge is that acoustic cues exhibit many-to-many relations with symptoms and are sensitive to confounders (age, sex, channel conditions), requiring a structured mapping that remains efficient for edge execution while producing interpretable outputs.

This paper presents an approach that links acoustic features to DSM‑5 \cite{americanpsychiatricassociation2022dsm5} depression indicators through a transparent specification that runs locally, preserving privacy while enabling near‑real‑time analysis. The contributions are:
\begin{itemize}[leftmargin=*,topsep=2pt,itemsep=1pt]
  \item A \textbf{Linkage Framework} that maps acoustic features to DSM-5 depressive-behavior indicators through explicit, testable rules.
  \item An \textbf{edge-first design} with a privacy-preserving pipeline demonstrating real-time throughput on commodity HW.
  \item \textbf{Preliminary evaluation} on DAIC-WOZ \cite{usc_daic_woz} showing directionally consistent feature–indicator associations.
\end{itemize}

\section{System Design}
\label{sec:design}

\begin{figure*}[t]
    \centering
    \includegraphics[width=\textwidth]{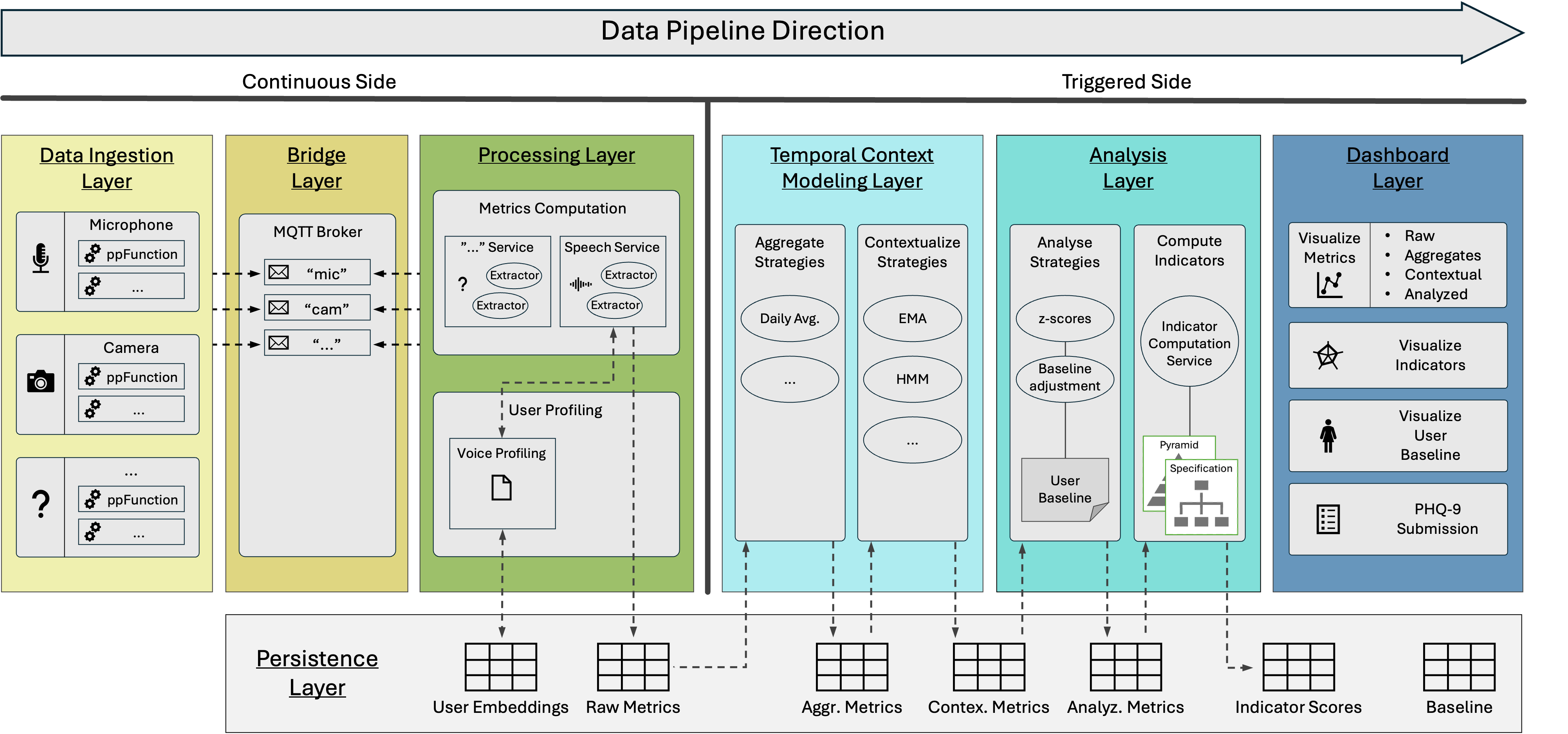}
    \caption{High-level system architecture showing the data pipeline from audio ingestion to DSM-5 indicator scoring.}
    \label{fig:high-level-architecture}
\end{figure*}

Moving beyond aggregate depression scores to symptom-level understanding is essential for appropriate treatment \cite{fried2025depressionsumscores}. While deep learning achieves high accuracy, it often lacks the interpretability needed for clinical workflows \cite{herm2023mlexplainability}. Our approach prioritizes explainability through a structured pipeline (Figure~\ref{fig:high-level-architecture}).

The architecture comprises: (1)~a \textbf{data ingestion layer} that captures audio via microphones and applies VAD (SileroVAD \cite{SileroVAD}) to isolate speech; (2)~a \textbf{processing layer} extracting low-level descriptors ($F_0$, intensity, jitter, shimmer) on 25\,ms frames, aggregated into high-level descriptors over 10\,s windows; (3)~a \textbf{temporal context layer} applying indicator-specific smoothing; and (4)~an \textbf{analysis layer} mapping features to DSM-5 indicators. All processing above ingestion runs on-device, preserving privacy.

\subsection{Linkage Framework (LF)}

The LF bridges raw audio with DSM-5 symptom indicators (Table~\ref{tab:dsm5-symptoms}) through a layered, clinically-grounded mapping. Unlike end-to-end models that learn opaque representations, the LF makes each transformation explicit and testable.

\textbf{Layer structure.} The framework comprises five layers: \emph{measurements} (frame-level $F_0$, intensity, jitter, shimmer) $\rightarrow$ \emph{features} (functionals like $F_0$ range, harmonics-to-noise ratio, spectral slope) $\rightarrow$ \emph{biomarkers} (monopitch, monoloudness, slowed speech, prolonged pauses) $\rightarrow$ \emph{indicators} (DSM-5 symptom items) $\rightarrow$ \emph{analysis} (counting rule). Each layer has defined inputs/outputs, enabling modularity: features can be extended without changing clinical logic.

\textbf{Feature categories.} Two main feature families are extracted. \emph{Prosodic features} ($F_0$ statistics, intensity range, speech/articulation rate, pause ratios) capture pitch, loudness, and timing patterns associated with monotony and slowed speech~\cite{Low2020keyacousticfeatures}. Depressed individuals often exhibit ``prosodic flattening,'' characterized by compressed pitch range and diminished loudness dynamics~\cite{cummins2015review}. \emph{Source/quality features} (jitter, shimmer, HNR, CPP) reflect phonatory stability and breathiness, linked to fatigue and emotional flattening. Elevated jitter and shimmer indicate vocal fold tension changes, while reduced HNR suggests breathiness consistent with psychomotor fatigue~\cite{cummins2015review}.

\textbf{Clinical grounding.} Each feature-indicator link is grounded in clinical literature. Reduced $F_0$ variability (monopitch) reflects affective flattening in depressed patients, historically characterized as ``low, slow, hesitant, and monotonous'' speech~\cite{cummins2015review}. Longer pauses and slower articulation rate indicate psychomotor retardation, \ie difficulty initiating and sustaining speech~\cite{Low2020keyacousticfeatures}. Increased jitter and shimmer reflect reduced phonatory control associated with fatigue~\cite{Donaghy2024VoiceBiomarkers}. These associations are directional: reduced variability increases indicator scores, as does increased pause duration.

\begin{figure}[t]
    \centering
    \includegraphics[width=1.0\linewidth]{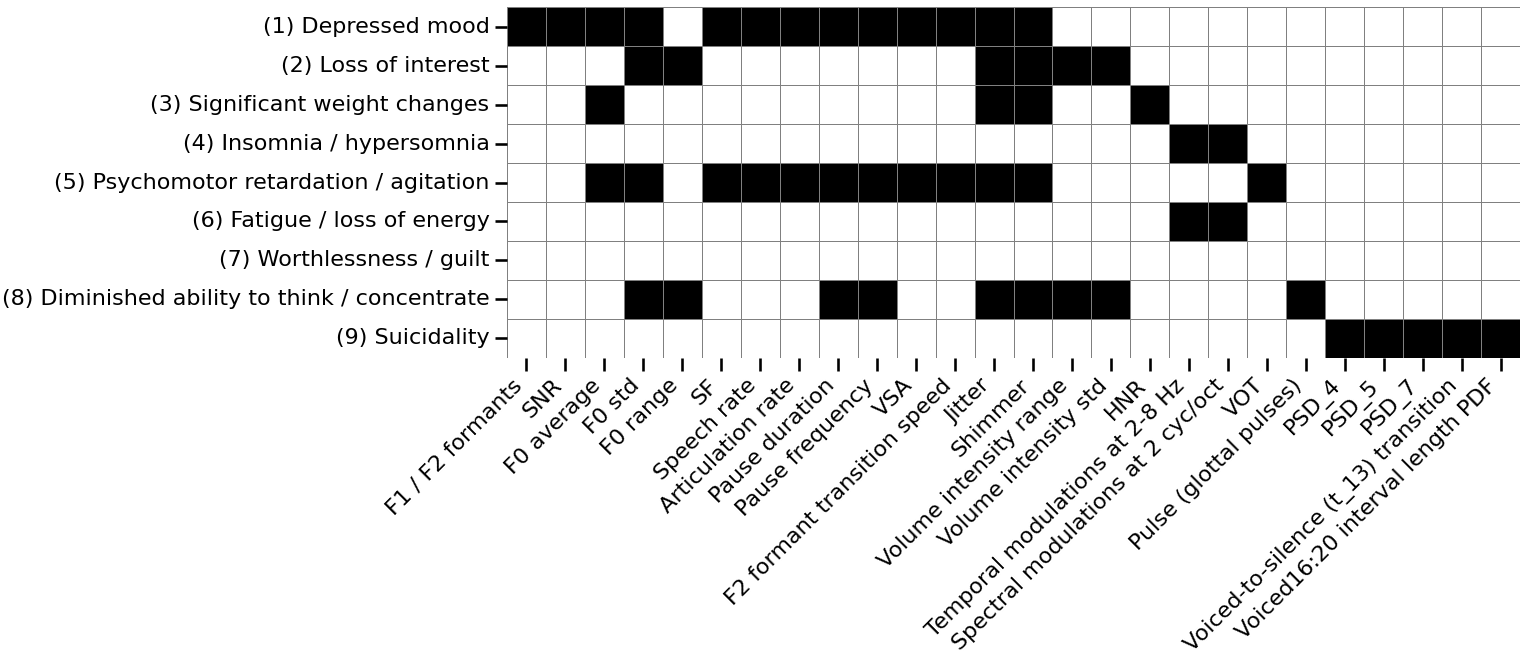}
    \caption{Feature-indicator intersections, showing many-to-many mappings.}
    \label{fig:feature_intersections_2}
\end{figure}

\textbf{Many-to-many mappings.} Crucially, mappings are \textbf{many-to-many}: one feature like $F_0$ range contributes to multiple indicators (\eg depressed mood \emph{and} loss of interest), while each indicator draws from multiple features (Figure~\ref{fig:feature_intersections_2}). This deliberate redundancy increases robustness against noise and missing data. For example, psychomotor retardation (indicator 5) is supported by pause duration, speech rate, articulation rate, and $F_0$ variability; if one feature is unreliable due to channel noise, others provide compensating evidence.

\begin{table}[t]
\centering
\small
\caption{DSM-5 symptom indicators for MDD \cite{americanpsychiatricassociation2022dsm5}.}
\label{tab:dsm5-symptoms}
\begin{tabular}{cl}
\toprule
\textbf{ID} & \textbf{Symptom Indicator} \\
\midrule
(1) & Depressed mood \\
(2) & Loss of interest or pleasure \\
(5) & Psychomotor retardation or agitation \\
(8) & Diminished ability to think/concentrate \\
\bottomrule
\end{tabular}
\begin{flushleft}
\footnotesize \textit{Note:} Full 9 indicators in DSM-5; those most observable from speech.
\end{flushleft}
\end{table}

\subsection{Computing Indicator Scores}

The transformation from features to DSM-5 indicator scores follows four steps. First, features are standardized via z-score normalization $z_m(t)=(x_m(t)-\mu_m)/\sigma_m$, with outliers clipped at $\pm 3\sigma$. Second, directional transforms align feature deviations with indicator semantics (\eg reduced pitch $\rightarrow$ higher depression score). Third, an exponential moving average (EMA) provides temporal stability:
\begin{equation}
\label{eq:ema}
\bar S_i(t)=(1-\beta_i)S_i(t)+\beta_i\,\bar S_i(t-1),
\end{equation}
where $\beta_i$ controls the response per indicator. Finally, the DSM-5 decision rule is applied: let $B_i(t)=\mathbb{1}\{\bar S_i(t) \geq \vartheta_i\}$ indicate the presence of symptoms. The support signal is:
\begin{equation}
\label{eq:mdd}
\textsc{MDD\_support}(t)=\mathbb{1}\left\{\sum_{i=1}^9 B_i(t)\ge 5 \land (B_1(t) \lor B_2(t))\right\},
\end{equation}
requiring at least five indicators including depressed mood (1) or loss of interest (2), matching DSM-5 \cite{americanpsychiatricassociation2022dsm5}. The formulation is reproducible from a configuration file. Threshold values $\vartheta_i$ can be personalized through periodic PHQ-9 self-reports, enabling detection of \emph{within-person} changes rather than relying on population-level norms.

\section{Preliminary Evaluation}
\label{sec:evaluation}

A preliminary evaluation was conducted to assess whether the Linkage Framework produces directionally consistent feature--indicator associations and whether the pipeline achieves real-time performance on edges.

\subsection{Dataset and Hypotheses}

We used DAIC-WOZ \cite{usc_daic_woz}, which contains speech recordings with PHQ-8 responses that align with DSM-5 indicators. A balanced sample of 64 participants (50\% male, 50\% female) was processed with VAD and aggregated into 10\,s windows. We tested four hypotheses linking acoustic patterns to indicators (5) psychomotor change and (8) concentration difficulty:
\textbf{H1}~reduced pitch variability (negative association);
\textbf{H2}~longer pauses (positive);
\textbf{H3}~lower energy dynamics (negative);
\textbf{H4}~slower speech tempo (negative).

\subsection{Results}

\begin{table}[t]
\centering
\small
\caption{Preliminary hypothesis outcomes on DAIC-WOZ.}
\label{tab:hyp_outcomes}
\begin{tabular}{lllll}
\toprule
\textbf{H} & \textbf{Feature} & \textbf{Ind.} & \textbf{Dir.} & \textbf{Outcome} \\
\midrule
H1 & $F_0$ variability & (5),(8) & -- & supported \\
H2 & Pause dur./freq. & (5),(8) & + & supported \\
H3 & Energy dynamics & (5),(8) & -- & partial \\
H4 & Speech tempo & (5),(8) & -- & supported \\
\bottomrule
\end{tabular}
\end{table}

\begin{figure}[t]
    \centering
    \includegraphics[width=\linewidth]{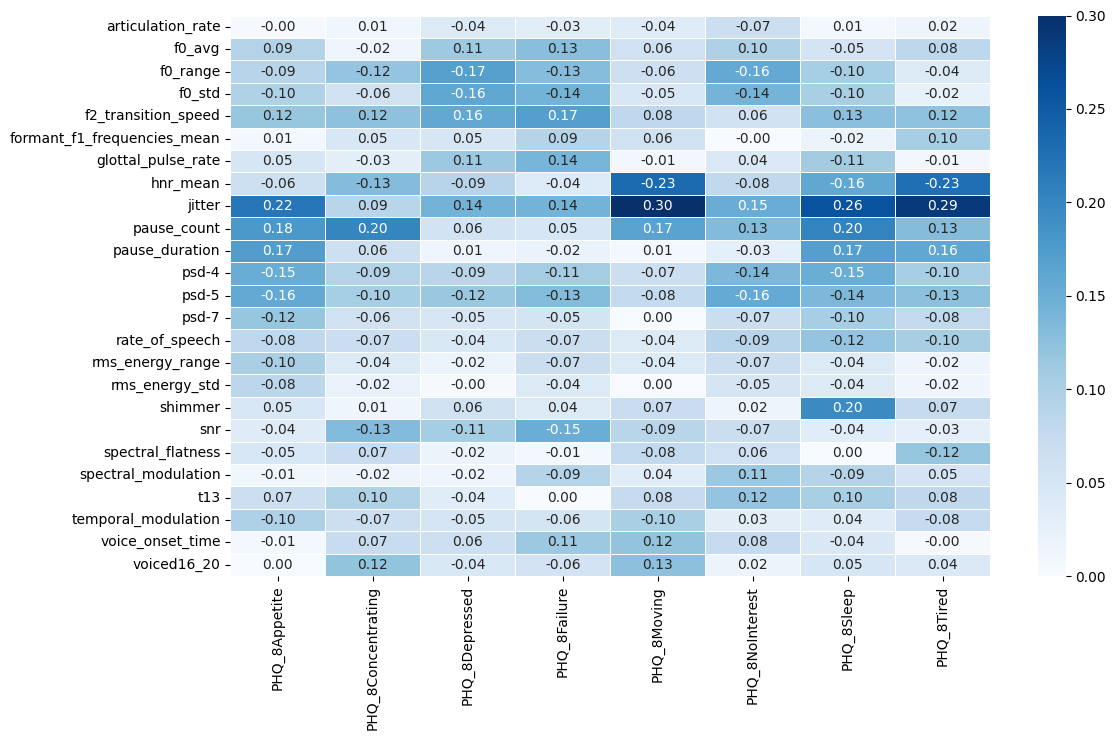}
    \caption{Effect sizes ($r$) for feature--indicator pairs (darker = stronger).}
    \label{fig:heatmap_r}
\end{figure}

Table~\ref{tab:hyp_outcomes} and Figure~\ref{fig:heatmap_r} summarize the preliminary findings. Reduced pitch variability (H1), longer pauses (H2), and slower speech tempo (H4) show directionally consistent associations with the hypothesized indicators. Energy dynamics (H3) show weaker effects. While effect sizes remain moderate and do not reach significance after FDR correction (consistent with the multifactorial nature of depression and DAIC-WOZ's limited sample size), the directional consistency supports the LF design.

\subsection{Temporal Behavior}

To assess streaming feasibility, we concatenated TESS~\cite{TESS_dataset} samples into synthetic audio alternating between depressed and non-depressed segments. The LF correctly accumulates sustained patterns while damping transient spikes via EMA smoothing. Brief emotional fluctuations do not trigger false positives; the temporal smoothing requires consistent patterns over multiple windows before indicator scores rise. This aligns with DSM-5's two-week persistence requirement.

\subsection{Edge Performance}

The pipeline was benchmarked on a MacBook Pro M1 (16\,GB RAM) under continuous audio load from TESS. After initial warm-up (~2 segments), processing sustained real-time throughput with latency below 1\,s per 10\,s window, confirming edge feasibility without GPU acceleration. A web-based dashboard provides interpretable views of raw metrics, contextualized features, and indicator scores; source code is publicly available~\cite{depressiondetection2025GitHubRepo}.

\subsection{Limitations}

Several limitations constrain interpretation. DAIC-WOZ provides single-session, interview-style speech, not longitudinal home recordings with naturalistic acoustic conditions. PHQ labels are self-reported proxies introducing potential recall bias. The sample size (n=64) limits statistical power; effect sizes show directional consistency but remain moderate after FDR correction, consistent with the multifactorial etiology of depression. The current LF covers four of nine DSM-5 indicators; symptoms like appetite change and sleep disturbance require complementary modalities. Despite these constraints, directional consistency across multiple feature families supports the design rationale.

\section{Summary and Future Work}
\label{sec:summary}

This paper introduced a framework~\cite{depressiondetection2025GitHubRepo} linking acoustic features to DSM-5 indicators through an explicit Linkage Framework. Unlike black-box models, our approach exposes which features support which indicators, enabling symptom-level review. The edge-first design ensures raw speech never leaves the home, addressing privacy concerns for household monitoring.

Preliminary evaluation on DAIC-WOZ shows directionally consistent associations for psychomotor change and concentration difficulty. Streaming evaluation confirms appropriate temporal behavior, damping transients while accumulating sustained patterns.

\textbf{Future work} includes: (1)~longitudinal validation in home environments with naturalistic speech; (2)~multimodal fusion with wearable signals for indicators not observable from speech; (3)~population calibration across languages and demographics; and (4)~clinical integration for screening workflows. The open-source implementation enables reproducibility toward transparent mental health monitoring.

\bibliographystyle{IEEEtran}
\balance
\bibliography{references.bib}

\end{document}